\pdfoutput=1
\documentclass[11pt]{article}

\usepackage[final]{acl}

\usepackage{times}
\usepackage{latexsym}
\usepackage{booktabs} % For better table styling
\usepackage{siunitx}  % For aligning decimal points
\usepackage{booktabs} % For professional-looking tables
\usepackage{multirow} % For multi-row cells
\usepackage{enumitem}
\usepackage{linguex}
\PassOptionsToPackage{x11names,dvipsnames}{xcolor}
\usepackage{xcolor}   % For cell highlighting
\definecolor{bluebell}{rgb}{0.64, 0.64, 0.82}
\definecolor{ceil}{rgb}{0.57, 0.63, 0.81}
\definecolor{apricot}{rgb}{0.98, 0.81, 0.69}
\usepackage[T1]{fontenc}
\usepackage[utf8]{inputenc}
\usepackage{subcaption}
\usepackage[nopatch=footnote]{microtype}
\usepackage{inconsolata}

\usepackage{graphicx}

\usepackage{amssymb}

\title{Expect the Unexpected? Testing the Surprisal of Salient Entities}

\author{Jessica Lin \and Amir Zeldes\\
         Department of Linguistics \\ Georgetown University \\ \texttt{\{yl1290, amir.zeldes\}@georgetown.edu}}
\begin{document}

\maketitle
\begin{abstract}
Previous work examining the Uniform Information Density (UID) hypothesis has shown that while information as measured by surprisal metrics is distributed more or less evenly across documents overall, local discrepancies can arise due to functional pressures corresponding to syntactic and discourse structural constraints. However, work thus far has largely disregarded the relative salience of discourse participants. We fill this gap by studying how overall salience of entities in discourse  relates to surprisal using 70K manually annotated mentions across 16 genres of English and a novel minimal-pair prompting method. Our results show that globally salient entities exhibit significantly higher surprisal than non-salient ones, even controlling for position, length, and nesting confounds. Moreover, salient entities systematically reduce surprisal for surrounding content when used as prompts, enhancing document-level predictability. This effect varies by genre, appearing strongest in topic-coherent texts and weakest in conversational contexts. Our findings refine the UID competing pressures framework by identifying global entity salience as a mechanism shaping information distribution in discourse.
\end{abstract}

\section{Introduction}

A large body of work in discourse processing has examined what makes certain entities central to a text. From a discourse perspective, texts invariably have central participants (entities the discourse is fundamentally ``about'') and peripheral ones that play supporting roles (\citealt{givon1983}, \citealt[182--185]{Chafe1994}). This notion of centrality operates at the document level, differing from local attentional salience studied at the utterance-level e.g.~in Centering Theory \cite{grosz-etal-1995-centering}. While local salience determines which entities are most prominent at any given moment through mechanisms such as grammatical role and recency, global salience captures which entities are most important to the discourse as a whole. Little is known about how this global notion of salience interacts with information distribution across extended discourse.

Information-theoretical work suggests that speakers tend to structure text such that information flow is approximately even \cite{fenk1980konstanz,NIPS2006_c6a01432}. This Uniform Information Density (UID) hypothesis posits that surprisal (negative log-probability) of new words should remain relatively constant. However, perfect uniformity is hardly achieved because multiple competing pressures simultaneously shape how speakers and writers build discourse \cite{clark-etal-2023-cross, pimentel-etal-2021-disambiguatory, tsipidi-etal-2024-surprise}. Recent work has identified syntactic constraints \cite{clark2022russian}, phonological factors \cite{pimentel-etal-2021-disambiguatory}, and discourse structures \cite{tsipidi-etal-2024-surprise} as pressures that create departures from uniformity. This raises the question: does discourse-level referential structure also exert pressure on information distribution? 
%If globally salient entities provide readers with stronger expectations about what kinds of information are likely to come next-for instance, mentioning ``discrimination'' in a paper about the prevalence of racial discrimination makes the following discourse more predictable than peripheral entities do. In this view, global discourse salience constitutes a discourse-structural pressure that interacts with, and sometimes competes with, the tendency toward uniform information density.
On the one hand, salient entities may be high in informational content, and therefore more surprising. On the other hand, they may be central to the topics of the documents they appear in, making document content more predictable when they appear. Rather than predicting a global shift in average information density, we hypothesize that global salience gives rise to localized non-uniformities in discourse. In particular, salient entities tend to be followed by stretches of reduced surprisal, creating troughs in surprisal contours that reflect increased expectations about upcoming content.

To investigate this relationship, we leverage a dataset of over 70,000 entity mentions across 16 genres of spoken and written English, in which entities are annotated for global salience based on summary-worthiness \cite{lin-zeldes-2025-gum}: entities appearing consistently across multiple independent summaries of a document are treated as globally salient. We examine how this measure relates to surprisal computed using language model probabilities, addressing three research questions:

\begin{enumerate}
    \item Do globally salient entities themselves exhibit different surprisal profiles than non-salient entities in naturally occurring discourse?
    \item Do globally salient entities systematically reduce surprisal for surrounding discourse content compared to non-salient entities?
    \item Does the relationship between global salience and surprisal vary systematically across genres?
\end{enumerate}

For RQ1, we find significant differences in raw corpus data where multiple pressures operate simultaneously. For RQ2, we develop a minimal-pair paradigm controlling for confounds to demonstrate that globally salient entities increase predictability for document contents. For RQ3, we find stronger effects in more topic-coherent genres (i.e.~ones focuses on single topic, such as academic articles) than in topic-shifting ones (e.g.~conversation).

Our results provide evidence, to our knowledge for the first time, that globally salient entities are more surprising and systematically reduce surprisal for upcoming content. For UID, our results support a view in which uniformity is a tendency shaped by competing constraints, with global salience contributing a referential mechanism that interacts with predictability across discourse.

\section{Related Work}

\subsection{UID and Competing Pressures} The Uniform Information Density (UID) hypothesis proposes that speakers tend to distribute information so that surprisal remains approximately even across a discourse \cite{fenk1980konstanz, NIPS2006_c6a01432}. A large body of work supports this tendency, showing that speakers avoid sequences of extremely high or low surprisal when alternative formulations are available. For example, speakers use optional complementizers (``that'') more often before less predictable clauses to smooth information flow \cite{NIPS2006_c6a01432}.

However, many studies have shown systematic departures from uniformity driven by independent linguistic pressures. \citet{pimentel-etal-2021-disambiguatory} report a robust cross-linguistic pattern in which word-initial segments carry higher surprisal than word-final segments, reflecting phonotactic and information structural pressures to front-load disambiguatory information even when this produces locally uneven information distribution. At the syntactic level, \citet{clark-etal-2023-cross} demonstrate that real word orders are more uniform than most counterfactual reorderings, but that only linguistically implausible grammars achieve perfectly uniform profiles, indicating that grammatical constraints limit the degree of uniformity that can be achieved. For longer texts, \citet{tsipidi-etal-2024-surprise} show that surprisal varies systematically across a document and that hierarchical discourse structure predicts these non-uniform contours. Together, these findings motivate viewing UID as a tendency that interacts with other linguistic and discourse pressures to shape information distribution.

\subsection{Discourse Salience and Surprisal}
\label{sec:dis_sal_surp}
The notion of salience in discourse processing operates at multiple levels. Local salience, studied extensively in Centering Theory, determines which entities are most prominent at any given moment through mechanisms such as grammatical role, recency, and pronominalization \cite{grosz-etal-1995-centering}. By contrast, global salience captures document-level importance: which entities the discourse is fundamentally ``about'' \cite{givon1983,grosz-sidner-1986-attention}. These views are complementary rather than competing, addressing different aspects of salience.

Previous work on whether salient entities themselves are more or less predictable has yielded mixed results. Some studies suggest that marked or salient forms carry higher surprisal \cite{Racz2013, ZarconeEtAl2016SalSurprisal}, while others show that discourse-prominent entities become more predictable through repeated mention and establish topicality \cite{givon1983, arnold2010speakers}. Multiple factors influence entity predictability, including grammatical role \cite{grosz-etal-1995-centering}, recency \cite{arnold2025does}, and referential form choice \cite{arnold2010speakers, rohde2014grammatical}, making it difficult to isolate salience effects in natural contexts.

No previous study has examined the surprisal of globally salient entities, which have been operationalized through summary-worthiness in previous work \cite{lin-zeldes-2025-gum}. We hypothesize that globally salient entities are more surprising locally, but reduce surprisal across extended discourse for other document contents, creating systematic departures from uniform information distribution, analogous to phonotactic or syntactic pressures identified in prior work.

\subsection{Operationalizing Global Entity Salience} While there are many kinds and definitions of salience (see \citealt{HeusingerSchumacher2019,BoswijkColer2020}), work in Computational Linguistics has focused on the identification of the most prominent, notable or memorable entities in a document, which can be identified by a number of methods, including direct annotation via human intuition \cite{dojchinovski2016crowdsourced}, extraction from user click-stream data \cite{gamon2013identifying}, use of hyperlinks and categories in Wikipedia data \cite{wu2020wn} and automatic alignment with document summaries \cite{dunietz2014new}. The latter paradigm in particular has recently been extended to the extraction of gradient salience ratings based on multiple summaries of a document \cite{lin-zeldes-2025-gum}, based on the idea that if an entity is salient it will be difficult to write a summary without mentioning it, and that the number of summaries mentioning an entity can represent its relative salience. Below we use this definition and data to collect and compare surprisal and salience scores.

\section{Methods}\label{sec:methods}

\subsection{Data}
The dataset we use for this paper, \textbf{GUM}-based \textbf{S}ummary \textbf{A}ligned \textbf{G}raded \textbf{E}ntities (GUM-SAGE, \citealt{lin-zeldes-2025-gum}), is built on version 11 of the GUM corpus \cite{zeldes2017gum}, an open-access, multilayer resource for English spanning over 250K tokens across 16 genres. GUM features a variety of annotations such as Universal Dependencies parses \cite{de-marneffe-etal-2021-universal}, entity types, Wikification \cite{lin-zeldes-2021-wikigum}, discourse parses \cite{liu-etal-2024-gdtb}, and crucially for our purposes, coreference resolution \cite{zhu-etal-2021-ontogum,Zeldes2022}. 

Each document in GUM is accompanied by 5 summaries following strict guidelines \cite{liu-zeldes-2023-gumsum}, which are either all human-written (in the dev/test sets) or one human and 4 model-generated ones (training set). In GUM-SAGE, entities mentioned in the documents were aligned to the summaries to derive salience scores corresponding to \textbf{summary worthiness}, based on the idea that summaries will tend to include the most salient entities. A score of 5 is assigned to entities present in all summaries, and a score of 1 to entities appearing in only one, with 0 representing totally non-salient entities (mentioned in no summary, about 84.5\% of entities in the dataset). 

Since the data contains manual coreference annotations, we are able to rate all mentions\footnote{We use \textit{mention} to refer to any textual span annotated as referring to an entity in GUM, including proper names, common nouns, and pronouns. GUM permits nested mentions, so phrases such as \ref{ex:aardvarks} [the tips of the noses of the aardvarks] contain multiple markables (e.g., the aardvarks, the noses of the aardvarks, the tips of the noses of the aardvarks), each treated as a distinct entity mention.} of each entity, regardless of how it is mentioned (e.g.~proper name, common noun or pronoun). The data contains just over 70K mentions of over 31K unique entities. See Table~\ref{tb:genre_overview_salience} for an overview of the data.

\begin{table*}[t]
\centering
\resizebox{0.8\textwidth}{!}{%
\begin{tabular}{lrrrrrrrrr}
\toprule
\textbf{Genre} & \textbf{Docs} & \textbf{Tokens} & \textbf{Modality} & \textbf{Mentions} & \textbf{Entities} & \textbf{\% Top1} & \textbf{\% Top2} & \textbf{\% Top3} & \textbf{\% Salient} \\
\midrule
Academic writing      & 18 & 17,169 & Written & 5,055 & 3,001 & 1.90 & 3.35 & 5.00 & 15.65 \\
Biographies           & 20 & 18,213 & Written & 5,772 & 3,052 & 2.32 & 3.71 & 5.57 & 16.02 \\
Vlog                  & 15 & 16,864 & Spoken  & 4,499 & 1,256 & 1.63 & 3.44 & 5.61 & 14.83 \\
Conversations         & 15 & 17,932 & Spoken  & 4,531 & 1,166 & 1.12 & 2.49 & 4.64 & 16.24 \\
Courtroom transcripts & 9  & 11,148 & Spoken  & 2,938 & 1,053 & 1.80 & 3.46 & 5.82 & 14.68 \\
Essays                & 9  & 10,842 & Written & 3,061 & 1,672 & 0.93 & 1.74 & 2.89 & 10.88 \\
Fiction               & 19 & 17,511 & Written & 4,977 & 2,018 & 2.51 & 3.55 & 4.96 & 13.53 \\
Forum (reddit)        & 18 & 16,364 & Written & 4,544 & 1,958 & 2.10 & 2.71 & 3.95 & 10.23 \\
How-to guides         & 19 & 17,081 & Written & 4,468 & 2,011 & 2.17 & 2.85 & 4.22 & 14.73 \\
Interviews            & 19 & 18,196 & Spoken  & 5,216 & 2,293 & 2.98 & 3.49 & 4.21 & 6.68 \\
Letters               & 12 & 9,989  & Written & 2,848 & 1,325 & 3.34 & 5.22 & 7.10 & 19.62 \\
News stories          & 24 & 17,186 & Written & 5,023 & 2,305 & 3.80 & 5.03 & 7.40 & 15.56 \\
Podcasts              & 10 & 11,986 & Spoken  & 3,059 & 1,163 & 1.46 & 2.62 & 4.37 & 11.37 \\
Political speeches    & 15 & 16,720 & Spoken  & 4,847 & 2,297 & 1.84 & 2.71 & 3.46 & 7.79 \\
Textbooks             & 15 & 16,693 & Written & 4,719 & 2,687 & 1.55 & 2.59 & 3.31 & 8.27 \\
Travel guides         & 18 & 16,515 & Written & 4,471 & 2,559 & 1.09 & 1.45 & 2.54 & 10.63 \\
\midrule
\textbf{Total GUM}    & \textbf{255} & \textbf{250,409} & - & \textbf{70,028} & \textbf{31,816} & \textbf{2.06} & \textbf{3.18} & \textbf{4.71} & \textbf{13.21} \\
\bottomrule
\end{tabular}%
}
\caption{Overview of GUM-SAGE genres with document counts, token counts, modality, mention and entity counts, and proportion of salient entities by salience level (Top1: score=5, Top2: score $\geqslant$ 4, Top3: score $\geqslant$ 3). Percentages are calculated over all entities per genre.}
\label{tb:genre_overview_salience}
\end{table*}

\subsection{Surprisal}
\label{sec: method-surprisal}
% I took these out because we have little space in a short paper and we'll basically need to re-state all of this in the sections of the experiments - AZ

%To investigate the relationship between entity salience and surprisal, we formulated two hypotheses and designed a series of experiments to test them while controlling for confounding factors.

%\textbf{Hypothesis 1}: Because salient entities are more relevant to the overall document meaning, they are expected to facilitate the prediction of subsequent content. As a result, sentences following \textbf{salient entities} should exhibit \textbf{lower surprisal} than those following non-salient entities.

%\textbf{Hypothesis 2}: The relationship between salience and surprisal is expected to become clearer after controlling for confounding factors such as entity position, length, and information status. We hypothesize that salient entities, being central to the discourse, will become more predictable and therefore exhibit lower surprisal—particularly in later parts of the text.

We measured entity surprisal using different experimental scenarios to systematically control for potential confounding factors (see below), but in all cases, token-level surprisal scores\footnote{Token-level surprisal is the standard unit in UID research, since autoregressive language models define probabilities incrementally over tokens. Higher-level units (e.g., mentions or sentences) necessarily require aggregation from token-level estimates \cite{NIPS2006_c6a01432, meister-etal-2021-revisiting, pimentel-etal-2021-disambiguatory, clark2022russian}} were computed by processing each document with a sliding window of 1,024 tokens using a distilled version of \textsc{GPT-2} language model (See Appendix~\ref{sec:appendix-exp-setup} for implementation details). For a multi-word mention such as \textit{a full romantic evening}, mention-level surprisal scores were computed by averaging token-level surprisals, and entity-level surprisal scores for the cluster of all mentions of a single entity in a document were obtained by averaging mention-level scores across all mentions of the entity, based on the gold standard coreference annotations available in the corpus. 

Since surprisal can be affected by document length (the longer the document, the more context for later mentions), and document lengths vary across genres, we apply standard scaling to raw mean surprisal scores, resulting in \textbf{mean surprisal z-scores} (i.e.~an entity's mentions' mean surprisal score can be \textit{z} standard deviations above or below the mean of its document). We use this method throughout the paper, and when surprisal scores are discussed they should be understood in terms of negative and positive standard deviations away from the mean scores for mentions or entities in the document, as appropriate. This procedure allows us to normalize across potentially dissimilar documents, which are inevitably rather different in a dataset spanning a broad range of spoken and written genres, and also makes intuitive sense if we want to contrast the salient and non-salient entities in each document as possibly positive or negative deviations from an average baseline.

%In Experiment 2, prompts consist of full entity mentions. In Experiment 3, we use only the syntactic head noun extracted from UD parses to further control for length and nesting effects.

\section{Experiments}
\label{sec: experiments}
We present three experiments examining how global salience relates to surprisal. Experiment 1 (Section~\ref{sec:exp1}) measures entity-level surprisal in natural contexts, establishing a baseline that reveals how multiple pressures operate simultaneously in discourse. Experiments 2 and 3 (Section~\ref{sec:exp2} and \ref{sec:exp3}) focus on a complementary hypothesis, that salient entities will reduce the surprisal of their documents' contents more when used as prompts. Here we propose a novel minimal-pair paradigm to control for position, length, and nesting confounds, allowing us to measure whether globally salient entities reduce surprisal regardless of where they appear.

Since we aim to examine whether salience-predictability relationships generalize across diverse discourse contexts, all experiments use the GUM-SAGE dataset described in Section~\ref{sec:methods}, which provides gradient salience scores across 16 genres of spoken and written English.

\subsection{Natural context analysis}\label{sec:exp1}

We first test whether globally salient entities exhibit different surprisal profiles than non-salient entities by measuring average token-level surprisal per mention of each entity and correlating this with salience scores.

\begin{figure}[h!tbp]
  \includegraphics[width=\columnwidth]{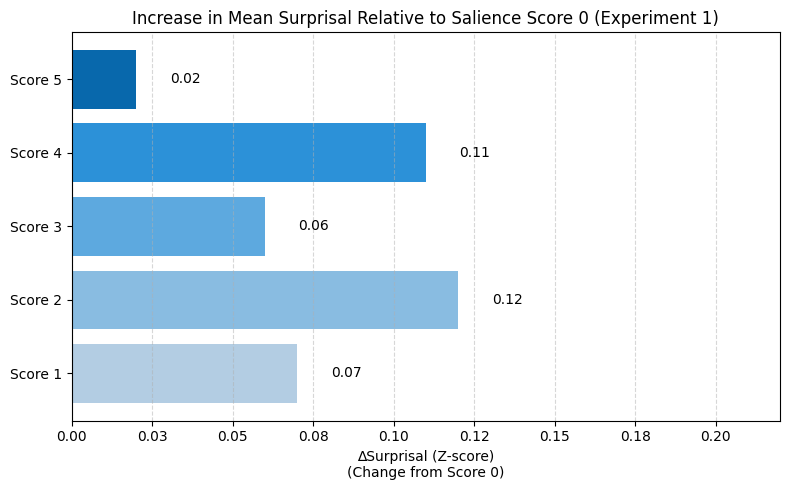}
  \caption{Change in mean surprisal for salience scores 1–5 relative to score 0 in Experiment 1.}
  \label{fig:box_graded_exp1}
\end{figure}

Figure~\ref{fig:box_graded_exp1} shows mean surprisal changes relative to entities with salience score 0. Salient entities (scores 1--5) show significantly higher surprisal than non-salient entities ($t =-2.15$, $p =0.031$), though the effect size is small ($\sim$0.1 standard deviations), and varies seemingly at random with salience score. This demonstrates that discourse-level salience (or `summary-worthiness') exerts a measurable but almost negligible influence on information distribution even in naturally occurring text, though salience score matters less.

%Figure~\ref{fig:genre_exp1} examines binary salience (score > 0 vs. = 0) across genres, revealing erratic patterns: salient entities in academic papers show higher surprisal than non-salient ones, while biography articles show the opposite pattern by a wide margin.

%\begin{figure}[htbp]
%  \includegraphics[width=\columnwidth]{latex/images/genre_exp1_1.png}
%  \caption{Mean surprisal scores across genres for Experiment 1 (salience score > 0 in orange, score=0 in green).}
%  \label{fig:genre_exp1}
%\end{figure}
%\vspace{-6pt}

\begin{table*}[t]
\centering
\small
\setlength{\tabcolsep}{6pt}
\begin{tabular}{lrrrr}
\toprule
Predictor & Coef. & $P$-value & 95\% CI & $\Delta$AIC \\
\midrule
Intercept          & 0.8205  & *** & [0.764, 0.877] & -- \\
Position           & 0.0002  & *   & [0.000, 0.000] & 2.68 \\
Salience score & -0.0471 & *** & [-0.070, -0.025] & 14.86 \\
Mention length       & -0.0535 & *** & [-0.059, -0.048] & 352.11 \\
Is Nesting                 & -0.8493 & *** & [-0.903, -0.796] & 886.27 \\
\bottomrule
\end{tabular}
\caption{Results from an OLS regression model \texttt{Surprisal \textasciitilde{} Salience score + Mention length + Is Nested + Position}, with predictors ordered by $\Delta$AIC from single-term deletions (least to most informative): larger $\Delta$AIC indicates a larger degradation in model fit when that predictor is removed. Significance codes: $^{*}p<0.05$, $^{**}p<0.01$, $^{***}p<0.001$.}
\label{tb:regression}
\end{table*}

To understand why the effect is so weak, we consider several confounding factors that complicate the relationship between salience and surprisal, using a linear regression model shown in Table~\ref{tb:regression}. 

%Table~\ref{tb:regression} shows that all predictors are significant ($p<0.05$), highlighting the confounded nature of surprisal in natural context. Nesting has the strongest effect ($\beta=-0.849$, $\Delta$AIC=886.27), followed by mention length ($\beta=-0.054$, $\Delta$AIC=352.11); the negative coefficients indicate that more deeply nested entities and longer mentions are associated with lower surprisal. Salience remains significant when controlling for these factors ($\beta=-0.047$, $\Delta$AIC=14.86), consistent with prior work showing that discourse-prominent entities become more predictable through repeated mention and topicality \cite{givon1983, arnold2010speakers}. Position contributes minimally but remains statistically significant ($\beta=0.0002$, $p=0.031$, $\Delta$AIC=2.68). These confounding effects operate as follows:

\textbf{Position effects.} Position shows a positive effect ($\beta=0.0002$, $p<0.05$, $\Delta$AIC=2.68), indicating higher surprisal for mentions later in documents, perhaps due to less explicit descriptions.

\textbf{Length and function words.} Mention length substantially reduces surprisal ($\beta=-0.054$, $p<0.001$, $\Delta$AIC=352.11). Longer mentions are more predictable on average because averaging token-level surprisal disproportionately weights highly predictable function words. For example, out of context, \ref{ex:aardvarks} has lower mean surprisal than \ref{ex:people} despite containing the improbable noun \emph{aardvarks}, driven by repeated predictable tokens such as \emph{the} and \emph{of}. 

\ex. [The tips of the noses of the aardvarks]$_{5.23}$\label{ex:aardvarks}

\ex. [People]$_{8.12}$\label{ex:people}

\indent \textbf{Nesting.} 
Nesting shows the strongest effect ($\beta=-0.849$, $p<0.001$, $\Delta$AIC=886.27): mentions that nest other mentions exhibit lower surprisal. This occurs because internal context makes the head noun more predictable. For example, comparing [the aardvarks] with [the noses of the aardvarks], the latter has lower average surprisal because ``noses of'' provides syntactic context that constrains what follows, whereas transitioning from external content to an NP is less predictable.

With these confounds incorporated, salience becomes a much more significant predictor, confirming that discourse salience influences information distribution substantially, with +1 salience score comparable to an extra word in length ($\sim$-0.05 coefficient for both). However, to establish a two way connection between salience and surprisal, we would also like to test whether salient entities are higher in content in the sense that they can make surrounding text less surprising, to which we turn next.

%However, the presence of multiple competing pressures motivates us to examine whether salient entities are focal points controlled minimal-pair paradigm in Experiments 2--3, where we isolate salience effects more directly. 

\subsection{Surprisal reduction from salient entities}  
\label{sec:exp2}

While measuring the surprisal of entity mentions in running texts is straightforward, measuring how informative an entity is for its text is more complicated: firstly, because some of the text in a document can precede that entity, and secondly, because of the confounds identified above: non-salient entities nesting salient ones may appear to reduce surprisal like the salient entities they contain, and longer mentions will provide more context, creating an unfair comparison. In order to test whether the content of salient entities reduces surprisal more than non-salient entities in a controllable way, we therefore develop two minimal-pair paradigms. The main idea of both is to test identical sentences paired with either salient or non-salient entity mention prompts.

In the first minimal-pair design, for each entity in a document we extract its first non-pronominal mention and place it as a prompt followed by a colon. We then measure the mean surprisal of each sentence in the document when placed after this prompt. For example, in a document about discrimination which mentions that psychologists have studied the phenomenon, we create pairs like:

\ex. \textbf{Discrimination:} \textit{The prevalence of discrimination across racial groups in contemporary America.}\label{ex:prompt1}

\ex. \textbf{Psychologists:} \textit{The prevalence of discrimination across racial groups in contemporary America.}\label{ex:prompt2}

Here, ``Discrimination'' (salient, score = 5) and ``Psychologists'' (non-salient, score = 0) from the same document are both followed by identical sentence content. We then compute surprisal only for the sentence tokens (excluding the prompt and colon), then average across all sentences in the document to obtain each entity's discourse-level effect on facilitating the predictability of its document's contents.

Because sentence lengths vary, for cross-document comparison, we standardize sentence-level scores into z-scores using document-level means (${\mu}_{doc}$) and standard deviations (${\sigma}_{doc}$) of entity-level averages (the mean of sentence surprisal scores $\overline{Surp}(sentence)$):

\begin{equation}
z=\frac{(\overline{Surp}(sentence)-{\mu}{doc})}{{\sigma}{doc}}
\end{equation}

This formulation naturally controls for position effects since we test each entity against all sentences in its document. If globally salient entities establish discourse-wide predictability, then on average they should reduce mean surprisal more than non-salient entities when used as prompts. However, we cannot control for length or nesting using this method, issues we address in Section~\ref{sec:exp3}.

\begin{figure}[h!tbp]
  \includegraphics[width=\columnwidth]{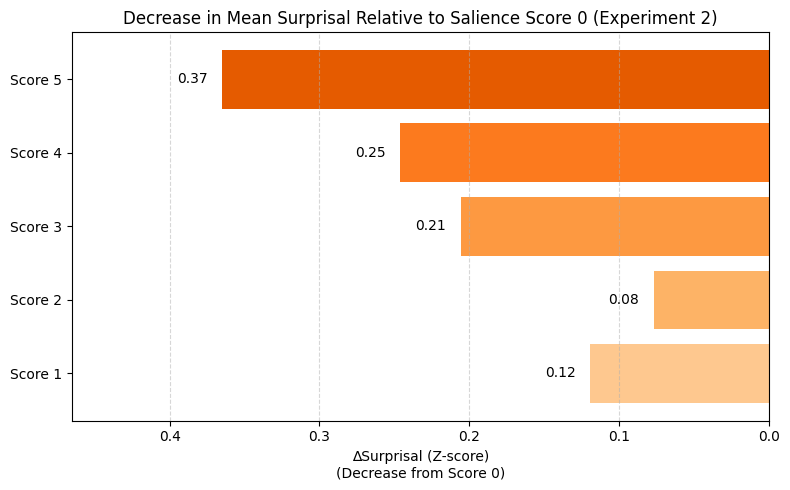}
  \caption{Decrease in mean surprisal relative to salience score 0 in Experiment 2. Magnitude of surprisal reduction largely grows with salience, indicating that more salient entities facilitate sentence-level predictability.}
  \label{fig:box_graded_exp2}
\end{figure}

\begin{figure}[h!tbp]
  \includegraphics[width=\columnwidth]{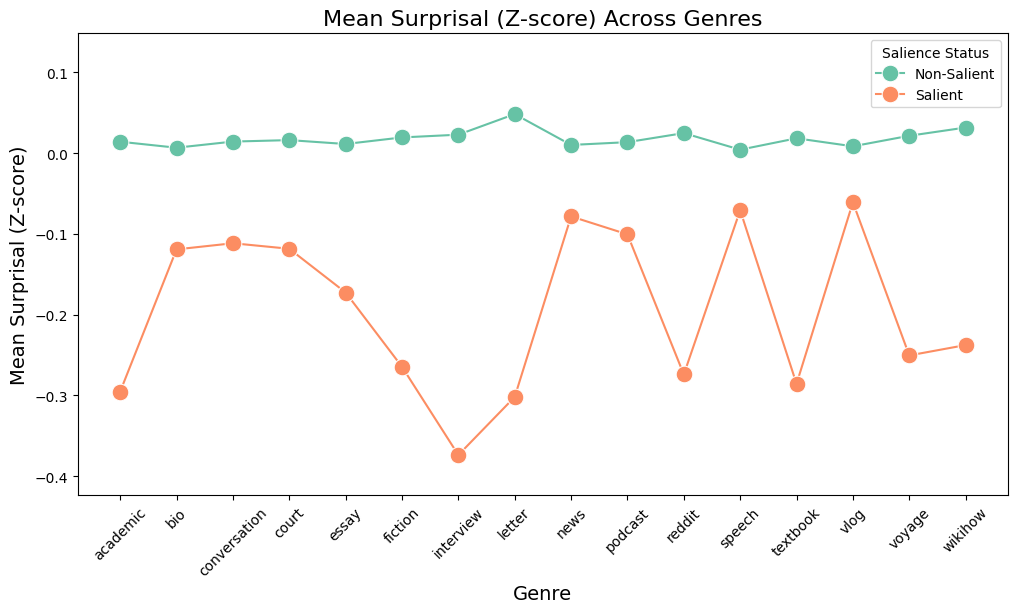}
  \caption{Mean surprisal scores across genres for Experiment 2 (salience score>0 in orange, score=0 in green).}
  \label{fig:genre_exp2}
\end{figure}

Figures~\ref{fig:box_graded_exp2} and \ref{fig:genre_exp2} show results for this setup. Figure~\ref{fig:box_graded_exp2} presents mean surprisal decreases relative to score 0 entities. All salience levels show negative values, indicating that salient entities reduce sentence surprisal when used as prompts. The pattern is nearly monotonic: higher salience corresponds to lower average z-scored surprisal, with sentences following non-salient entity prompts remaining close to baseline ($0.0171$ in Experiment 2; $0.0369$ in Experiment 3), while those following highly salient entities (score = 5) show substantially lower values ($-0.3481$ in Experiment 2; $-0.4710$ in Experiment 3).
%This contrasts sharply with Experiment 1's inconsistent patterns, demonstrating that controlling for position effects reveals clearer salience-predictability relationships.

Figure~\ref{fig:genre_exp2} shows consistent cross-genre patterns: salient entities yield lower mean surprisal than non-salient ones across all 16 genres (orange consistently below green). Maximum differences appear in formal interviews, scholarly writing (\texttt{academic} and \texttt{textbook}), informative texts (\texttt{voyage}, i.e.~travel guides from wikivoyage, and how-to guides from \texttt{wikihow}), with \texttt{speech} and \texttt{vlogs} much less pronounced. This universality across diverse discourse types suggests global salience robustly shapes predictability regardless of modality or register.

However, entity length and nesting remain confounds. Function word sequences like ``of the'' within multi-word mentions are highly predictable, potentially inflating the apparent predictability of longer entities. Additionally, non-salient entities can nest salient ones, diluting distinctions. Experiment 3 addresses these issues through head-noun isolation, at the cost of less accurate representations of entity phrases.

\subsection{Paired Head Noun Design}
\label{sec:exp3}
%RESULTS
To eliminate length and nesting confounds, Experiment 3 extracts syntactic head nouns from entities' first non-pronominal mentions using GUM's gold standard UD syntax tree annotations. The head token for each entity mention is assumed to be the token whose dependency parent token is outside of the span of the mention. Representing mentions using their head tokens eliminates common sub-sequences (e.g.,``of the''), standardizing prompt length to single tokens while removing predictable function words.\footnote{WordPiece tokenization may still create multiple tokens, but we expect impacts to be modest compared to phrase-level effects.} We apply the same minimal-pair design, measuring sentence surprisal following head-noun prompts.

\begin{figure}[htbp]
  \includegraphics[width=\columnwidth]{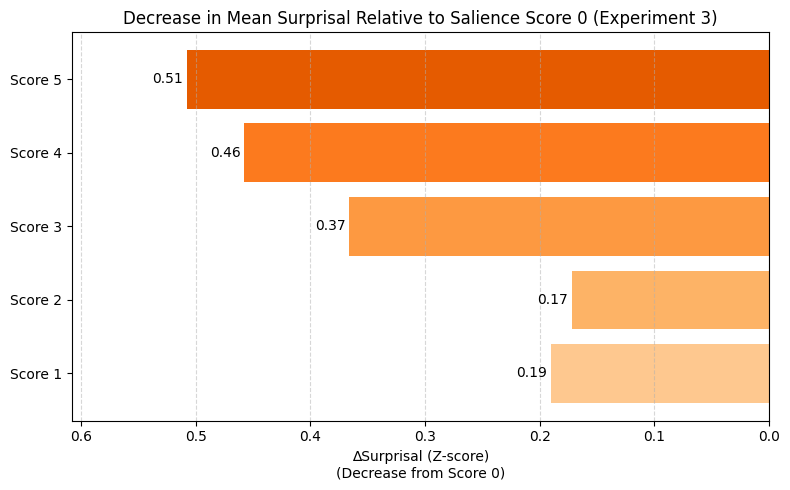}
  \caption{Decrease in mean surprisal relative to salience score 0 in Experiment 3.}
  \label{fig:box_graded_exp3}
\end{figure}

\begin{figure}[h!tbp]
  \includegraphics[width=\columnwidth]{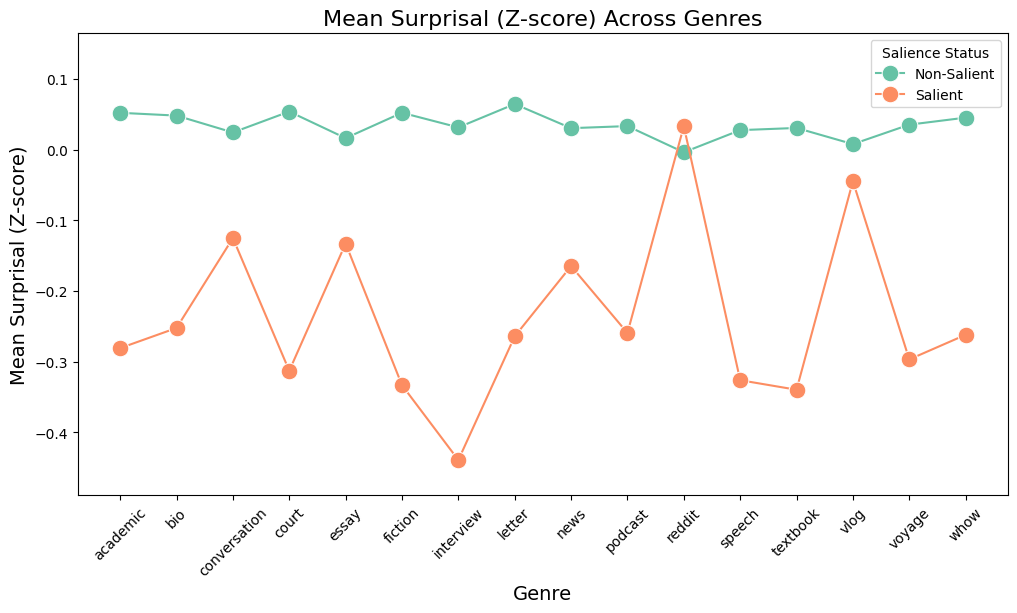}
  \caption{Mean surprisal scores across genres for Experiment 3 (salience > 0 in orange, salience=0 in green).}
  \label{fig:genre_exp3}
\end{figure}

Figure~\ref{fig:box_graded_exp3} shows stronger, steeper surprisal reduction as salience increases compared to Experiment 2. The minor misordering of scores 1 and 2 persists but is negligible. Genre patterns remain consistent (Figure~\ref{fig:genre_exp3}), with \texttt{interviews} leading, followed by \texttt{fiction}, \texttt{speeches}, and courtroom transcripts. \texttt{Reddit} and \texttt{vlogs} are borderline.

Compared to Experiment 2, this stricter design controls for both length and nesting. While head-noun reduction sacrifices semantic completeness, the persistence of robust salience-predictability relationships across both experiments demonstrates that global salience does interact with discourse predictability. These effects do not merely correlate with surface features like length or syntactic complexity, the potential confounds mentioned above. Instead, they reveal a discourse-structural pressure: globally salient entities increase expectations for upcoming content, yielding localized troughs in surprisal in the sentences that follow. These effects align with a view of UID in which systematic non-uniformities arise at specific points in discourse due to identifiable pressures analogous to phonotactic and syntactic pressures.

To verify that these patterns are not model-specific, we replicate all experiments using \textsc{GPT2-Small} (124M parameters), seeing similar trends; full figures and analyses are provided in Appendix ~\ref{sec:appendix-GPT2-small-replication}.

\section{Genre Analysis}\label{sec:genre}
The results from Experiment 3 suggest that globally salient entities reduce surprisal, but whether this effect persists across genres remains an open question. If global salience functions as a discourse-structural pressure competing with UID, its strength could vary with discourse structure itself. Genres differ systematically in discourse structure, topic continuity, and interactivity (especially for multi-party conversations). Expository writing (academic articles, textbooks) maintains tight topic focus and hierarchical organization, while conversational genres show frequent topic shifts (multi-party dialogue, forum discussions), lowering predictability. We therefore hypothesize that the salience effect should be strongest in topic-coherent genres and weakest in topic-shifting ones.

\subsection{Quantitative analysis}

\begin{figure*}[htbp]
  \centering
  \includegraphics[width=0.75\textwidth]{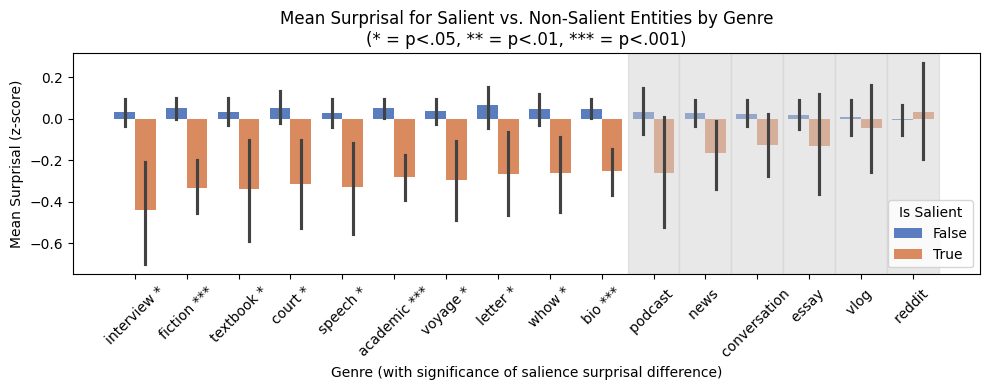}
  \caption{Mean surprisal (z-score) for salient vs. non-salient entities across genres. Bars show mean surprisal for salient (orange) and non-salient (blue) entities, with genres sorted by surprisal difference. Error bars represent 95\% confidence intervals. Asterisks indicate significant differences based on adjusted t-tests (*~$p$~<~.05,\ **~$p$~<~.01,\ ***~$p$~<~.001). Genres shaded gray show no significant difference between salient and non-salient entities.}
  \label{fig:genre_barplot}
\end{figure*}

Figure~\ref{fig:genre_barplot} shows mean surprisal z-scores for salient (orange) and non-salient (blue) entities across 16 genres using data from Experiment 3, sorted by surprisal difference. Error bars show 95\% confidence intervals adjusted for multiple testing, with asterisks for significant differences ($^{}p<.05$, $^{}p<.01$, $^{}p<.001$), and non-significant genres shaded gray.

In all genres but \texttt{reddit}, salient entities reduce surprisal more predictable than non-salient ones, and in most genres significantly so. This pattern is especially robust in \texttt{academic}, \texttt{fiction}, and \texttt{bio}, where differences are large and error bars do not overlap. These genres exhibit structured or goal-directed discourse supporting repeated, predictable use of salient referents (e.g.~protagonists, research topics or methods).

In contrast, the surprisal gap is smaller and non-significant in \texttt{conversation}, \texttt{vlog}, and inverted for \texttt{reddit}. These informal, topic-shifting genres show greater surprisal variability and overlapping confidence intervals. Salience aligns less closely with predictability here, suggesting genre moderates the salience-predictability relationship. Results for \texttt{essay} and \texttt{podcast} are also not significant, though these have substantially less data in the corpus (see Table \ref{tb:genre_overview_salience}).

These patterns reveal how competing pressures interact. In some genres, UID pressure toward evenness is systematically counteracted by discourse-structural pressure favoring predictability zones around central entities. In other more topic-shifting genres, topic instability may prevent salient entities from creating sustained predictability, allowing UID pressure to dominate more. The balance between these pressures can vary with discourse context and coherence structure.

\subsection{Qualitative Analysis}
\label{sec: qualitative_analysis}
\begin{figure*}[t!h]
\centering
\begin{subfigure}[b]{0.49\textwidth}
\includegraphics[width=\textwidth]{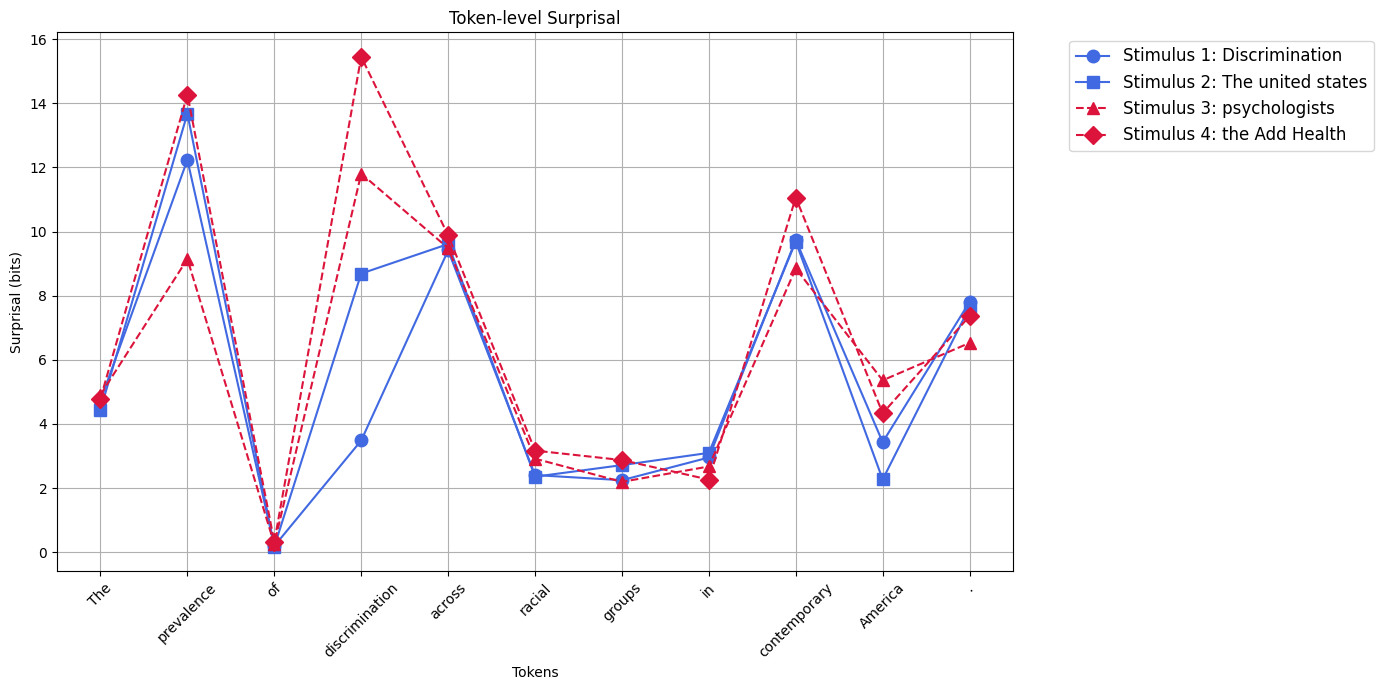}
\caption{Surprisal contour for a minimal pair in an academic excerpt.}
\label{fig:ex_ac}
\end{subfigure}
\hfill
\begin{subfigure}[b]{0.49\textwidth}
\includegraphics[width=\textwidth]{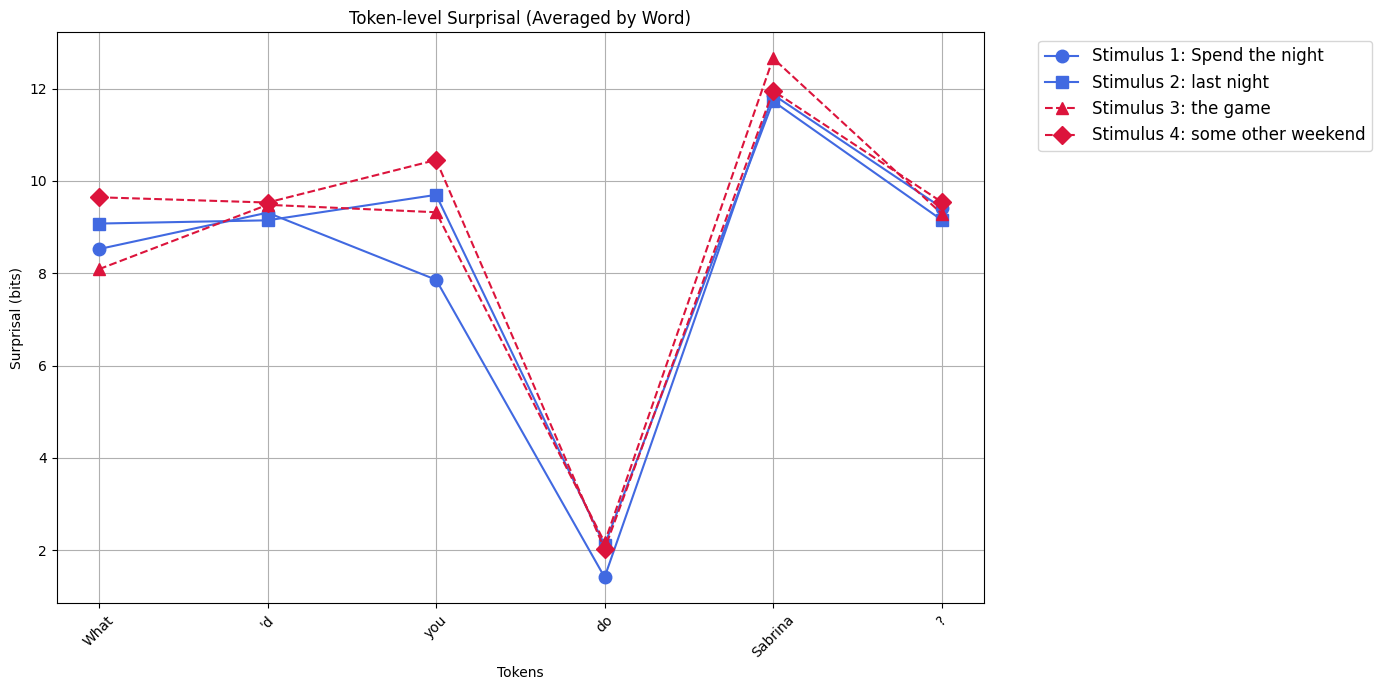}
\caption{Surprisal contour for a minimal pair in a conversation.}
\label{fig:ex_cn}
\end{subfigure}

\caption{Surprisal contour in different genres. Blue lines represent the surprisal contour of the sentence followed by salient entities; Red lines represent the surprisal contour of the sentence followed by non salient entities. Each stimulus is distinguished by different markers.}
\label{fig:genre_ex}
\end{figure*}

%upcoming content like
Figure~\ref{fig:genre_ex} illustrates the patterns above through contrasting examples. In an academic article on discrimination (Figure~\ref{fig:ex_ac}), salient entities like ``discrimination'' and ``the United States'' yield lower surprisal across sentences. These entities align with the discourse topic, constraining expectations and making upcoming content like ``racial groups'' or ``contemporary America'' more predictable than with non-salient ``psychologists'' or ``Add Health'' (the name of a study, which is less salient).

Conversely, in a conversation between ``Sabrina'' and her mom about staying somewhere overnight and other topics (Figure~\ref{fig:ex_cn}), salient entities like ``last night'' or ``Spend the night'' do not reliably reduce surprisal across the document. The sentence ``What'd you do, Sabrina?'', with low semantic overlap with the rest of the text, yields nearly identical surprisal curves for non-salient entity prompts like ``some other weekend''. This reflects looser topic structure: frequent topic shifts mean even salient content offers little predictive help.

These examples highlight why predictive utility of salient entities varies by genre. In structured expository texts with strong coherence, global salience creates robust predictability zones (systematic departures from UID driven by referential structure). In conversational settings with weak coherence, lexical cohesion is lower and the facilitatory effect diminishes, allowing information to distribute more evenly as UID would predict.

\section{Conclusion}
\label{sec:conclusion}
We examined whether global discourse salience, operationalized as summary-worthiness, functions as a systematic pressure on information distribution in discourse. Three experiments reveal nuanced interactions between salience and surprisal. Experiment 1 showed that globally salient entities exhibit significantly higher surprisal than non-salient ones in natural contexts. A linear model revealed that while multiple pressures (position, mention length, nesting) operate simultaneously, salience remains a significant predictor of surprisal even alongside these confounding factors. Experiments 2 and 3 used a novel minimal-pair paradigm to isolate salience effects by controlling confounds. Results show that globally salient entities systematically reduce surprisal for surrounding discourse content when used as prompts. This effect is robust across most genres but varies predictably with discourse structure: stronger in topic-coherent texts, weaker in topic-shifting conversational contexts.

These findings establish that global salience introduces localized non-uniformities in discourse. Like syntactic and phonotactic pressures identified in prior work \cite{clark-etal-2023-cross, pimentel-etal-2021-disambiguatory}, the discourse prominence of entities competes with the tendency toward uniformity, with salient entities enhancing predictability of surrounding content. This effect is sensitive to discourse structure: salience-driven predictability is stronger in topic-coherent genres than in conversational ones where topic shifts dominate.

For UID theory, our results show where and why non-uniformities arise: salient entities increase expectations for subsequent discourse, producing localized reductions in surprisal, while leaving document-level averages unchanged and preserving overall information density. For discourse theory, these findings validate that summary-worthiness captures document-central entities from an information theoretic perspective: entities independently selected for summaries constrain discourse-wide predictability. This relationship between summarization and predictability opens questions about how discourse centrality shapes both what speakers express and how listeners process information across extended discourse. 

Although this paper focused exclusively on entity mentions, future work may be able to tell us more about the ways in which surprisal relates to salience at large, and current work on annotating summary-based salience for propositions \cite{ZeldesEtAl2026} should allow for follow up studies using methods analogous to the ones proposed in this study. A further avenue of possible research relates to the extent to which our findings in textual data could have parallels in multimodal inputs, where work on UID is now beginning to disentangle pressures on uniformity when language is grounded in visual perception \cite{gay-etal-2026-information}. We leave these areas to future research and hope that our data and methodology will be fruitful for these and other research questions.
 
% Beyond teaching us about how salience shapes discourse-level predictability, these results also support the meaningfulness of graded salience scores obtained via alignment with summaries. Since the salience scores used in this paper were extracted entirely independently of surprisal, it is conversely a non-trivial finding that entities which lower surprisal more for each document's sentences turn out to be more likely to be selected for inclusion in summaries. This opens up new research questions about the relationship between summarization and the notions of salience and surprisal, the study of which we hope this research will promote.

\section*{Limitations}
This study is limited to English, a high-resource language with abundant training data for the language model used to compute surprisal. It remains an open question whether the observed patterns involving salience and surprisal hold in lower-resource languages, where language models may have weaker estimates of discourse structure and referential continuity, and morphosyntactic constraints differ. In addition, while we operationalized predictability using token-level surprisal estimates, other measures such as entropy or mutual information may capture different dimensions of uncertainty in discourse processing. Future work could explore how these alternative metrics relate to referential salience across languages and model architectures. The operationalization of salience using summaries is also vulnerable to subjectivity in the summarization process itself, though we expect that the use of multiple summaries helps to mitigate the effects of any single outlier summaries. Additionally, we do not separately analyze how global salience interacts with other referential factors such as definiteness, NP type, or grammatical role. While these factors influence local mention-level predictability (as discussed in Section~\ref{sec:dis_sal_surp}), our focus is on document-level salience effects that operate across all mention types. Future work should examine how global and local salience factors interact to shape information distribution at both mention and discourse levels. Finally, although we employed standard practices from previous studies to conform to and ensure comparability with previous work, we acknowledge that different tokenization strategies, especially in terms of word-piece tokenization, may also influence the results, a limitation we leave for future studies on surprisal using LM probability estimates in general.

%\section*{Acknowledgments}

\bibliography{anthology,custom}

\appendix

\section{Experimental Setup and Hyperparameters}
\label{sec:appendix-exp-setup}
\textbf{Model.} We computed token-level surprisal using \textsc{DistilGPT-2}, a distilled version of GPT-2 small. DistilGPT-2 contains 82M parameters and has been shown to preserve the core distributional properties of its larger counterpart \cite{sanh2019distilbert} while being computationally efficient.

\textbf{Implementation.} We used the Hugging Face \texttt{Transformers} library. Documents were processed with a sliding window of 1,024 tokens to maintain consistent context length. For each token, surprisal was calculated as the negative log-probability: 

$\text{surprisal}(w_i) = -\log P(w_i | w_1, \ldots, w_{i-1})$. 

The model was run in fp32 precision with all parameters at their default pretrained values without fine-tuning.

\textbf{Computation.} Inference was performed on a single NVIDIA A100 GPU with batch size of 16. Tokenization used automatic padding and truncation to handle variable-length inputs within each batch.

\textbf{Aggregation.} Token-level surprisals were averaged to obtain mention-level scores, and mention-level scores were averaged to obtain entity-level scores, as described in Section~\ref{sec: method-surprisal}.

\section{Replication with GPT2-Small}
\label{sec:appendix-GPT2-small-replication}
\begin{figure}[h!tbp]
  \includegraphics[width=\columnwidth]{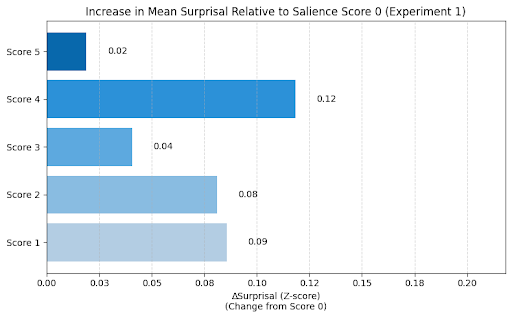}
  \caption{Change in mean surprisal for salience scores 1–5 relative to score 0 in Experiment 1 using \textsc{GPT2-Small}.}
  \label{fig:box_graded_exp1_gpt2}
\end{figure}

\begin{figure}[h]
  \includegraphics[width=\columnwidth]{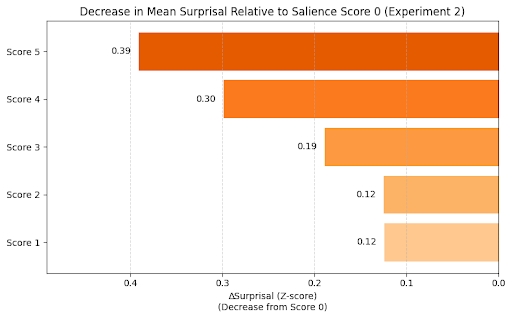}
  \caption{Change in mean surprisal for salience scores 1–5 relative to score 0 in Experiment 2 using \textsc{GPT2-Small}.}
  \label{fig:box_graded_exp2_gpt2}
\end{figure}

\begin{figure}[h]
  \includegraphics[width=\columnwidth]{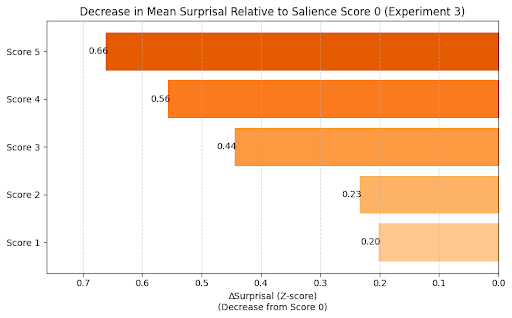}
  \caption{Change in mean surprisal for salience scores 1–5 relative to score 0 in Experiment 3 using \textsc{GPT2-Small}.}
  \label{fig:box_graded_exp3_gpt2}
\end{figure}

To assess whether the observed effects in Section~\ref{sec: experiments} depend on the choice of language model, we replicate all experiments using \textsc{GPT2-Small} (124M parameters). The results closely mirror those obtained with \textsc{DistilGPT-2} (82M parameters). In Experiment 1 (Figure~\ref{fig:box_graded_exp1_gpt2}), salient entities again exhibit slightly higher surprisal than non-salient ones. In Experiments 2 and 3 (Figures~\ref{fig:box_graded_exp2_gpt2}- \ref{fig:box_graded_exp3_gpt2}), we observe the same monotonic relationship between salience and surprisal reduction, with higher salience levels consistently yielding larger decreases (e.g., up to 0.66 in Experiment 3). The relative ordering across salience levels and the overall effect sizes remain comparable, indicating that our findings are not artifacts of a specific model architecture.

\vspace{-30pt}

\end{document}